\documentclass{article}

\usepackage{graphicx}
\usepackage{booktabs}
\usepackage{multirow}
\usepackage{algorithm}
\usepackage{algorithmic}
\usepackage{amsmath}
\usepackage{tabularx}
\usepackage{textcomp}  

\usepackage[dvipsnames]{xcolor}
\usepackage{colortbl}
\usepackage{array}
\usepackage{rotating} 
\usepackage{natbib}

\usepackage{graphicx} 
\usepackage{subcaption} 
\usepackage{CJKutf8}
\usepackage{enumitem}

\newcommand{\fancyname}{\text{SignCL}}

\usepackage{pifont} 
\usepackage{xcolor} 

\newcommand{\cmark}{\textcolor{green}{\ding{51}}} 
\newcommand{\xmark}{\textcolor{red}{\ding{55}}} 

\usepackage[final]{neurips_2024}

\usepackage[utf8]{inputenc} 
\usepackage[T1]{fontenc}    
\usepackage{hyperref}       
\usepackage{url}            
\usepackage{booktabs}       
\usepackage{amsfonts}       
\usepackage{nicefrac}       
\usepackage{microtype}      
\usepackage{xcolor}         

\title{Improving Gloss-free Sign Language Translation by Reducing Representation Density}

\author{
Jinhui Ye$^1$ \quad Xing Wang$^{4}$ \quad Wenxiang Jiao$^{4, \dagger}$ \quad Junwei Liang$^{1,2,\dagger}$ \quad Hui Xiong$^{1,2,3, \dagger}$ \\
$^1$Artificial Intelligence Thrust, HKUST (Guangzhou), Guangzhou, China\\
$^2$Department of Computer Science and Engineering, HKUST, Hong Kong SAR, China\\
$^3$Guangzhou HKUST Fok Ying Tung Research Institute\quad $^4$Tencent AI Lab  \\
\normalsize\tt jye624@connect.hkust-gz.edu.cn;  \normalsize\tt junweiliang@hkust-gz.edu.cn; \normalsize\tt  \\ xionghui@ust.hk; 
\normalsize\tt \{brightxwang,joelwxjiao\}@tencent.com \\
}

\begin{document}

\maketitle

\begin{abstract}
Gloss-free sign language translation (SLT) aims to develop well-performing SLT systems with no requirement for the costly gloss annotations, but currently still lags behind gloss-based approaches significantly.
In this paper, we identify \textbf{a representation density problem} that could be a bottleneck in restricting the performance of gloss-free SLT.
Specifically, the representation density problem describes that the visual representations of semantically distinct sign gestures tend to be closely packed together in feature space, which makes gloss-free methods struggle with distinguishing different sign gestures and suffer from a sharp performance drop. 
To address the representation density problem, we introduce a simple but effective contrastive learning strategy, namely {\fancyname}, which encourages gloss-free models to learn more discriminative feature representation
in a self-supervised manner.
Our experiments demonstrate that the proposed {\fancyname} can significantly reduce the representation density and improve performance across various translation frameworks. 
Specifically, {\fancyname} achieves a significant improvement in BLEU score for the Sign Language Transformer and GFSLT-VLP on the CSL-Daily dataset by 39\% and 46\%, respectively, without any increase of model parameters.
Compared to Sign2GPT, a state-of-the-art method based on large-scale pre-trained vision and language models, {\fancyname} achieves better performance with only 35\% of its parameters. Implementation and Checkpoints are available at \url{https://github.com/JinhuiYE/SignCL}.
\end{abstract}


\let\thefootnote\relax\footnotetext{$^\dagger$ Corresponding authors.}

\section{Introduction}
\label{sec:introduction}
Sign languages are the primary form of communication for millions of deaf individuals.  Sign language translation (SLT) aims to convert sign language into fluent spoken language sentences, which is a challenging task as it needs to extract information from continuous video and translate it into discrete text tokens. 
Most prior studies promoted the SLT by utilizing intermediate representations, namely gloss annotations, either directly or indirectly~\cite{camgoz2018neural, yin-read-2020-better,zhou2021improving, chen2022simple, zhangsltunet,ye-etal-2023-cross, wang2024degree}.
Gloss annotations are beneficial as they provide a simplified representation and sequential ordering of each gesture within continuous sign videos, which aids in representation learning for visual encoders. 
However, the creation of sign language translation datasets with gloss annotations is both resource-intensive and time-consuming. 

Recently, there has been a shift towards gloss-free sign language translation methods, which do not rely on gloss annotations to train SLT models. These methods usually rely on general datasets~\cite{yin2023gloss}, general pretraning strategy~\cite{zhou2023gloss}, or general large-scale foundation models~\cite{wong2024signgpt} to promote gloss-free SLT. However, there is a substantial gap between the sign language domain and the general domain~\cite{ye2022scaling, kan2022sign}. 
Models trained with general strategies or datasets often fail to capture the subtle differences in semantically distinct gestures, which are crucial for accurately understanding a specific sign language.
Therefore, the performance of gloss-free methods still significantly lags behind that of gloss-based approaches.

\begin{figure}[t!]
\centering
\includegraphics[width=0.85\textwidth]{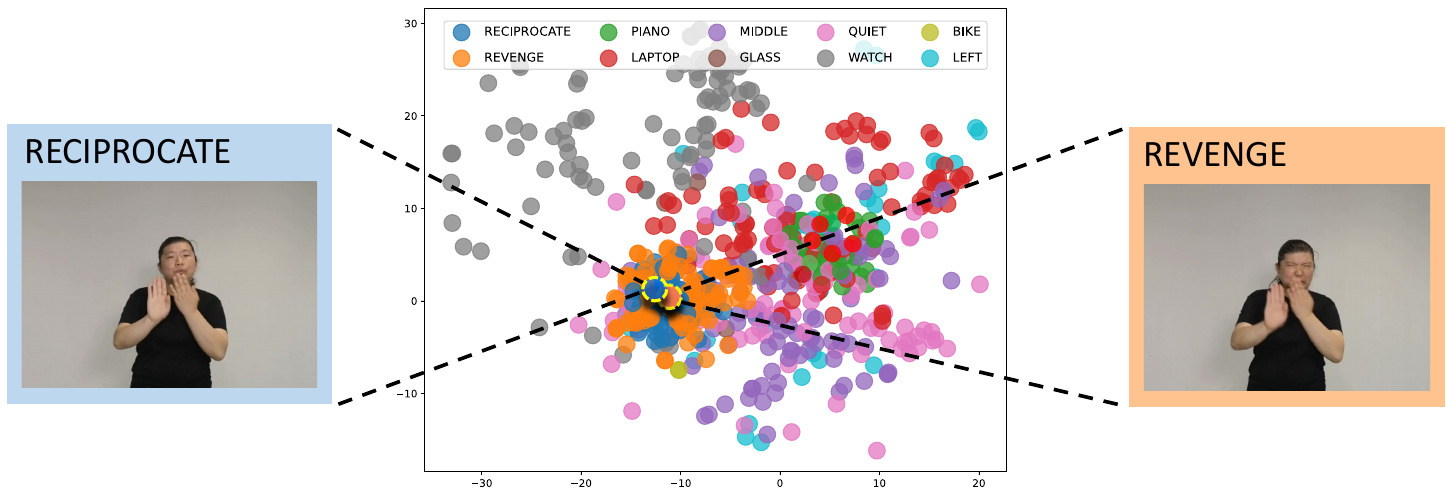}
\caption{An example of the representation density problem in sign language translation. The two images show the sign gestures for ``RECIPROCATE'' (blue dot) and ``REVENGE'' (orange dot). Although the two have opposite meanings, their visual representations are densely clustered together, as shown in the t-SNE visualization. The various colors in the visualization indicate sign gestures with different meanings.}
\vspace{-2 em}
\label{fig:motivation}
\end{figure}

In this paper, we identify \textbf{a representation density problem} in sign language translation: the visual representations of sign gestures with distinct semantics are likely to be close in representation space. 
This problem is attributed to the nature of sign language, a form of visual language that utilizes intricate hand gestures, facial expressions, and body movements to convey the signer‘s message~\cite{pizzuto2003review, vinson2010hands,guo2022context}. 
For example, in Figure~\ref{fig:motivation}, the signer performs sign gestures for opposite meanings, ``RECIPROCATE'' and ``REVENGE'', with similar visual information (i.e., only subtle differences in facial movements). The visual encoder in SLT models will encode similar visual information to visual representations in adjacent representation space, {even though they have distinct semantics}.
Without explicit gloss annotations, SLT models struggle to learn semantic boundaries in continuous sign videos and capture distinguishing visual representations for different sign gestures. As a result, the representation density problem poses a significant challenge for the SLT models in distinguishing between various sign gestures, leading to sharp performance drops. (Section ~\ref{sec:cSDR}). 

Further, we investigate various popular sign feature extraction methods, including gloss-based~\cite{camgoz2018neural, min2021visual} and gloss-free~\cite{yin2023gloss, zhou2023gloss}, to systematically study the representation density problem.  As shown in Figure~\ref{fig:sign-embeddings}, our investigation reveals that the representation density problem is prevalent across sign feature extraction methods.
Specifically, due to the lack of gloss annotations, the representation density problem appears to be more serious in gloss-free methods. Then, we conduct extensive SLT experiments and observe that  SLT models using gloss-free sign features as input consistently suffer a drop in performance in both sign language recognition and translation tasks compared to those using gloss-based sign features (Section ~\ref{sec:ImpactSDR}). 
Therefore, we demonstrate that the representation density problem can be a bottleneck in restricting the improvement of gloss-free sign language translation.

More importantly, we propose a simple but effective contrastive learning strategy named {\fancyname} to address the representation density problem. Specifically, {\fancyname} draws the visual representations of sign gestures with identical semantics closer together and pushes those with different semantics farther apart. Experimental results show that {\fancyname} can learn more distinctive feature representations and lead to significant improvements in terms of BLEU score on various well-known SLT frameworks (Section ~\ref{sec:main_exps}).
To summarize, the main contributions of this work are as follows:
\begin{itemize}[leftmargin=20pt]
    \item To the best of our knowledge, our work identifies the representation density problem in sign language translation for the first time. This problem is consistent across various sign feature extraction methods for SLT, including gloss-based and gloss-free methods.

    \item Experimental results empirically reveal that an increase in representation density leads to a significant performance drop in the accuracy of sign language recognition and translation. We find that the representation density problem poses a significant challenge for the gloss-free SLT.

    \item We propose a simple but effective contrastive learning strategy, namely {\fancyname}, to address the representation density problem. 
    Our experiments demonstrate that {\fancyname} can significantly enhance various well-known SLT frameworks.
    Specifically, {\fancyname} yields a 39\% BLEU score improvement for the Sign Language Transformer~\cite{camgoz2020multi} and a 46\% BLEU increase for GFSLT-VLP~\cite{zhou2021improving} on the CSL-Daily dataset.

\end{itemize}

\section{Related Works}

\subsection{Sign Language Translation}
\vspace{-0.4 em}
Sign Language Translation (SLT) methods can be broadly categorized into gloss-based and gloss-free approaches. For gloss-based methods, an essential factor is to directly or indirectly employ sign gloss annotations to improve sign video encoder performance \cite{camgoz2018neural, yin-read-2020-better, zhou2021improving, chen2022simple, zhangsltunet,chen2021semg}.  
These methods often employ Connectionist Temporal Classification~\cite{graves2006connectionist}  (CTC) loss to perform sign language recognition \cite{camgoz2020multi}.  Joint-SLT~\cite{camgoz2020multi} firstly introduces a multitask encoder-decoder framework with a CTCloss to soft-match sign representations and gloss sequences. STMC-T~\cite{zhou2021spatial} introducing intra-cue and inter-cue CTC loss to model multi-cue sequence information. 
Despite their effectiveness, creating SLT datasets with gloss annotations is resource-intensive and time-consuming. 
Gloss-free methods have emerged as a promising alternative, as they do not rely on gloss annotations during training, making them more generalizable. 
And recently, a growing body of literature has promoted the gloss-free SLT, such as GASLT~\cite{yin2023gloss} proposed local gloss attention to mimic gloss assistant, GFSLT~\cite{zhou2023gloss} adapted CLIP to do visual-language pretraining, and Sign2GPT~\cite{wong2024signgpt} promoted performance by making use of large-scale pre-trained vision and language models.
Nonetheless, the performance of gloss-free methods still significantly lags behind that of gloss-based approaches.

\vspace{-0.4 em}
\subsection{Contrastive Learning}
\vspace{-0.4 em}
Contrastive Learning~\cite{jaiswal2020survey, zhou2024mitigating, lin2024rethinking}, a popular unsupervised learning algorithm, aims to learn effective representations by pulling positive pairs closer together and pushing negative pairs farther apart. This approach has been widely utilized in both Natural Language Processing and Computer Vision~\cite{dai2023cloth2body}.
In Sign Language Translation (SLT), \citet{jin2021contrastive} utilize Contrastive Learning to create a Signer-Independent SLT model, using videos demonstrating signs from different signers as positive samples. Additionally, \citet{gan2023contrastive} proposes a visual-level contrastive learning method with various image augmentation strategies. ConSLT~\cite{fu2023token} do contrastive learning for effective token representation learning in text decoder. \citet{zhou2023gloss} and \citet{cheng2023cico} employ contrastive learning techniques to align video and text representations in SLT.
In this paper, we are the first one to address the representation density problem, focusing particularly on visual gesture duration as a central aspect.

\vspace{-0.4 em}
\subsection{Representation Density}
\vspace{-0.4 em}
Representation Density is often a focal point in classification tasks, also known as category density~\cite{bengio2013representation,xie2016unsupervised,schroff2015facenet,liu2016large, ma2021novel, zhang2023interactive,lin2022bert, entailtune:2024}.
This concept pertains to the compactness and clarity of feature representations across different categories.  
In the context of sign language, various methods have been developed to address the subtle nuances of sign actions. TSPNet~\cite{li2020tspnet} proposes a temporal hierarchical attention network to learn segmented representations.  HST-GNN~\cite{kan2022sign} utilizes a hierarchical spatio-temporal graph neural network to learn graph representations from multiple perspectives. GLE-Net~\cite{hu2021global} employs global contextual relationships and fine-grained cues to distinguish non-manual-aware features in isolated Sign Language Recognition. 
These methods are beneficial for addressing the subtleties of sign language movements. However, integrating them into existing state-of-the-art frameworks presents significant challenges, often resulting in performance disparities when compared to the SOTA. This paper is the first to propose the concept of representation density within this field and introduces {\fancyname}, which enhances the current mainstream transformer-based frameworks.

\section{Representation Density Problem}
\label{sec:signDensity}

This section investigates and identifies the representation density problem within existing sign feature extraction techniques, and examines whether representation density bottlenecks sign language recognition and translation performance.

\subsection{Preliminaries}

\paragraph{Existing Sign Feature Extraction Techniques}
Existing sign feature extraction methods can be divided into two categories: 1) gloss-based (e.g., Sign Recognition Pretrained~\cite{camgoz2018neural} and Self-Mutual Knowledge Distillation~\cite{min2021visual}) and 2) gloss-free (e.g., I3D Pretraining~\cite{yin2023gloss} and Visual-Language Pretraining~\cite{zhou2023gloss}). 
These methods were chosen for their representativeness in SLT and their well-documented open-source sign features. 

\begin{itemize}[leftmargin=20pt]
\item \textbf{Sign Recognition Pretrained (SRP)}~\cite{camgoz2018neural}: This approach leverages the sign language recognition datasets to train sign language recognition models and uses it as the feature extractor for the SLT task.
Notably, the features released by~\citet{camgoz2018neural} have been widely adopted as input features in a range of works~\cite{camgoz2020sign,zhou2021improving, jin2021contrastive, yao2023sign,ye-etal-2023-cross,chen2021semg2}.

\item \textbf{Self-Mutual Knowledge Distillation (SMKD)}~\cite{hao2021self}: This approach enhances SRP by enforcing the visual and contextual modules to focus on short-term and long-term information~\cite{hao2021self}. SMKD feature extraction has been shown to substantially enhance SLT translation performance compared to SRP~\cite{zhang2023sltunet, ye-etal-2023-cross}.

\item \textbf{I3D Pretraining (I3D)}~\cite{yin2023gloss}: This method employs I3D models as the backbone to pre-train the feature extractor, initially trained on the Kinetics dataset~\cite{kay2017kinetics} and subsequently fine-tuned on extensive web SLR datasets, such as WSLR~\cite{li2020word}.

\item \textbf{Visual-Language Pretraining (VLP)}~\cite{zhou2023gloss}: This method entirely forgoes gloss annotations and leverages a general visual-language pretraining strategy to align sign video representation with text. Embodied by GFSLT-VLP~\cite{zhou2023gloss}, this approach offers a more general solution that utilizes a broader range of sign language resources without the constraints of gloss annotations.

\end{itemize}

\paragraph{Representation Density Metrics}
Drawing inspiration from Fisher’s Discriminant Ratio (FDR)~\cite{8967450,harish2018hybrid}, a typical measure used to evaluate the discriminative power of features in the classification, we combine the average Inter-Gloss Distance and Intra-Gloss Distance into Sign Density Ratio (SDR, see Eqn.~\ref{eq:GDR}), which reflects the degree of representation density for each gloss $G_i$. This is given by the formula:

\begin{equation}
\label{eq:GDR}
SDR(G_i) = \frac{{D^{intra}_{G_i}} \,} {{avg. } D^{inter}_{G_i}}
=  \frac{D(G_i)}{{Mean}_{\substack{j \neq i}} \left( D(G_i, G_j) \right)}.
\end{equation}

Here, $D(G_i, G_j)$ represents the Inter-Gloss Distance between two glosses $G_i$ and $G_j$, and avg. $ {avg. } D^{inter}_{G_i}$ reflects the average distance of $G_i$ to all other glosses. The Intra-Gloss Distance ${D^{intra}_{G_i}}$ evaluates the average distance within a single gloss 
$G_i$. These distances are given by the following formulas:

\begin{equation}
\label{eq:Inter}
D(G_i, G_j) = \frac{1}{|G_i||G_j|} \sum_{x \in G_i, y \in G_j} d(x, y);
\end{equation}

\begin{equation}
\label{eq:Intra}
 D({G_i}) = \frac{1}{|G_i|(|G_i|-1)} \sum_{x, y \in G_i, x \neq y} d(x, y);
\end{equation}

Where, \( |G_i| \) and \( |G_j| \) denote the number of instances in glosses \( G_i \) and \( G_j \) respectively, and \( d(x, y) \) represents the distance measure between the embeddings of instances \( x \) and \( y \), i.e.,  euclidean distance.

The average Sign Density Ratio (SDR) of all glosses, denoted as ${SDR} = \text{Mean}(SDR(G_i))$, is calculated to evaluate the overall representation density of the dataset comprehensively.

\paragraph{Sign-Gloss Alignment}
To calculate the Sign Density Ratio (SDR), we need to determine the mapping relationship between input frames and gloss categories. Following previous works~\cite{kurzinger2020ctc, ye-etal-2023-cross}, we employ the CTC classifier as a sign-gloss forced aligner to establish the mapping between each gloss and its corresponding sign frames. The aligner provides the start position ${l}_v$ and end position ${r}_v$ within the video frame sequence for each corresponding gloss $g_v$. To optimize alignment performance on the test set, we merge the training and test datasets for comprehensive training and engage two volunteers to select the best frame $f_v$ from the range [$l_v$:$r_v$] to align with each gloss $g_v$. Extensive details on the training procedure and the aligner's performance metrics are documented in Appendix~\ref{tab:s2gAligner}.

\subsection{Demonstrating Representation Density Problem} 
\label{sec:cSDR}

\paragraph{Experiment Setups}
We primarily use the PHOENIX-2014T benchmark~\cite{camgoz2018neural} to investigate the representation density problem in existing sign feature extraction techniques. 
This benchmark was selected due to its rich collection of open-source sign features contributed by various research efforts.
We obtained the sign features by either downloading the officially released versions or reproducing the feature extraction process. 
Then, we employed t-SNE~\cite{van2008visualizing} to visualize the feature distribution of these semantically distinct sign gestures to investigate representation density.

\begin{figure}[htbp]
    \centering
    \hspace{1.8cm}  (a,b,c) Gloss-free Feature Extraction \hspace{1.0cm} (d,e) Gloss-based  Feature Extraction
    \\
    \begin{subfigure}{0.19\textwidth}
        \caption*{SDR=92.59\%}
        \includegraphics[width=\textwidth]{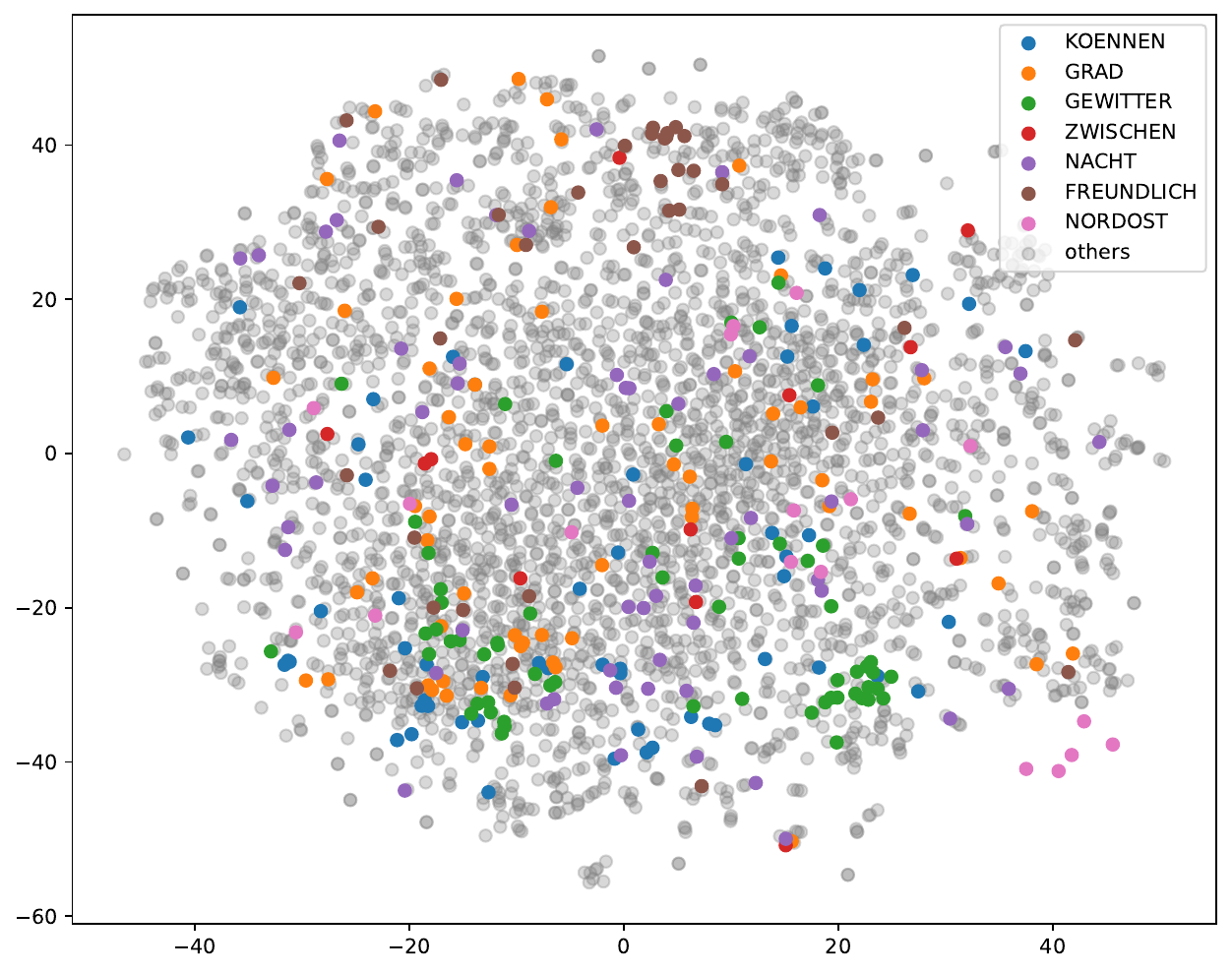}
        \caption{VLP}
        \label{fig:gfslt}
    \end{subfigure}
    \hfill 
    \begin{subfigure}{0.19\textwidth}
        \caption*{SDR=83.33\%}
        \includegraphics[width=\textwidth]{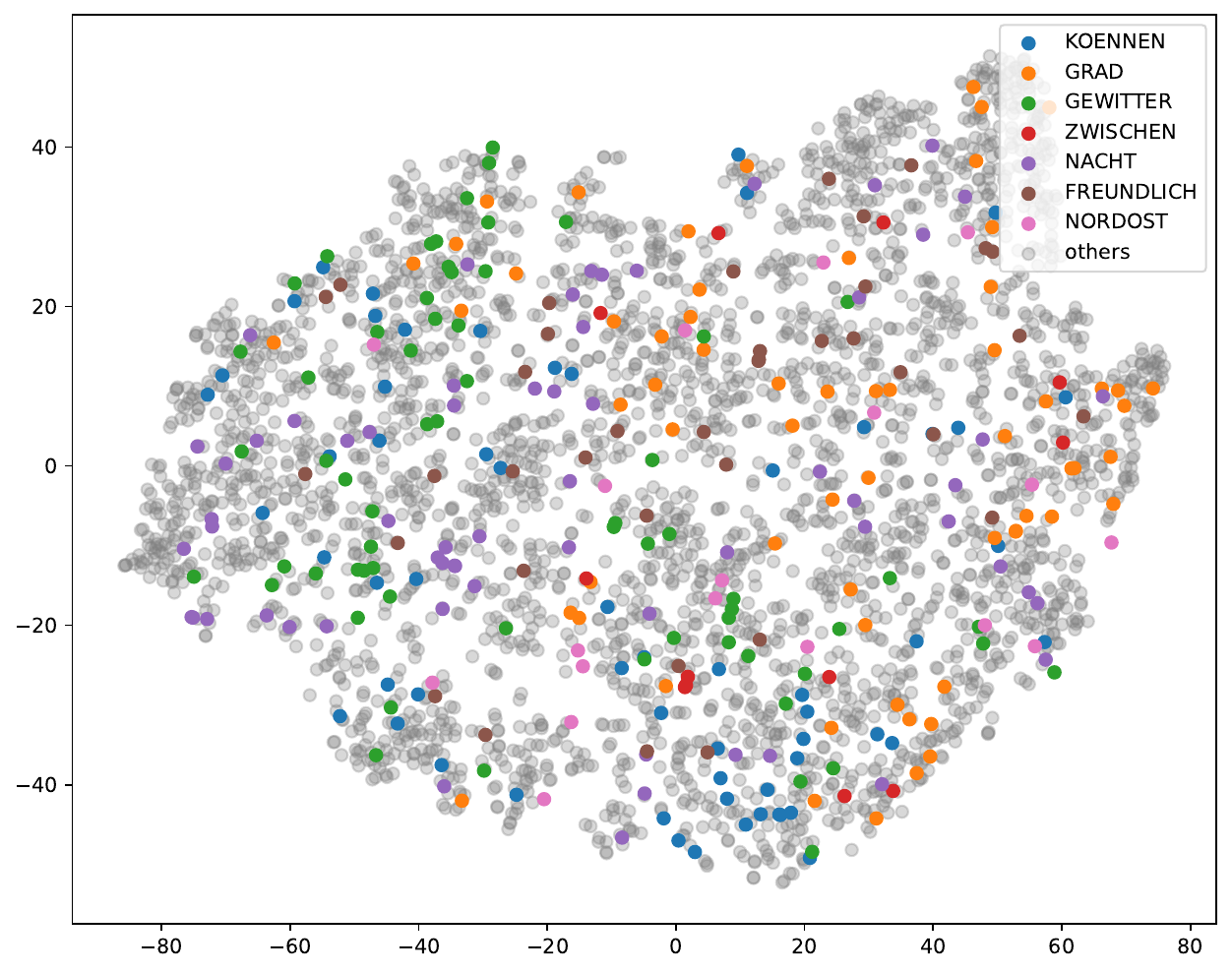}
        \caption{I3D}
        \label{fig:i3d}
    \end{subfigure}
    \hfill
    \begin{subfigure}{0.19\textwidth}
        \caption*{SDR=81.30\%}
        \includegraphics[width=\textwidth]{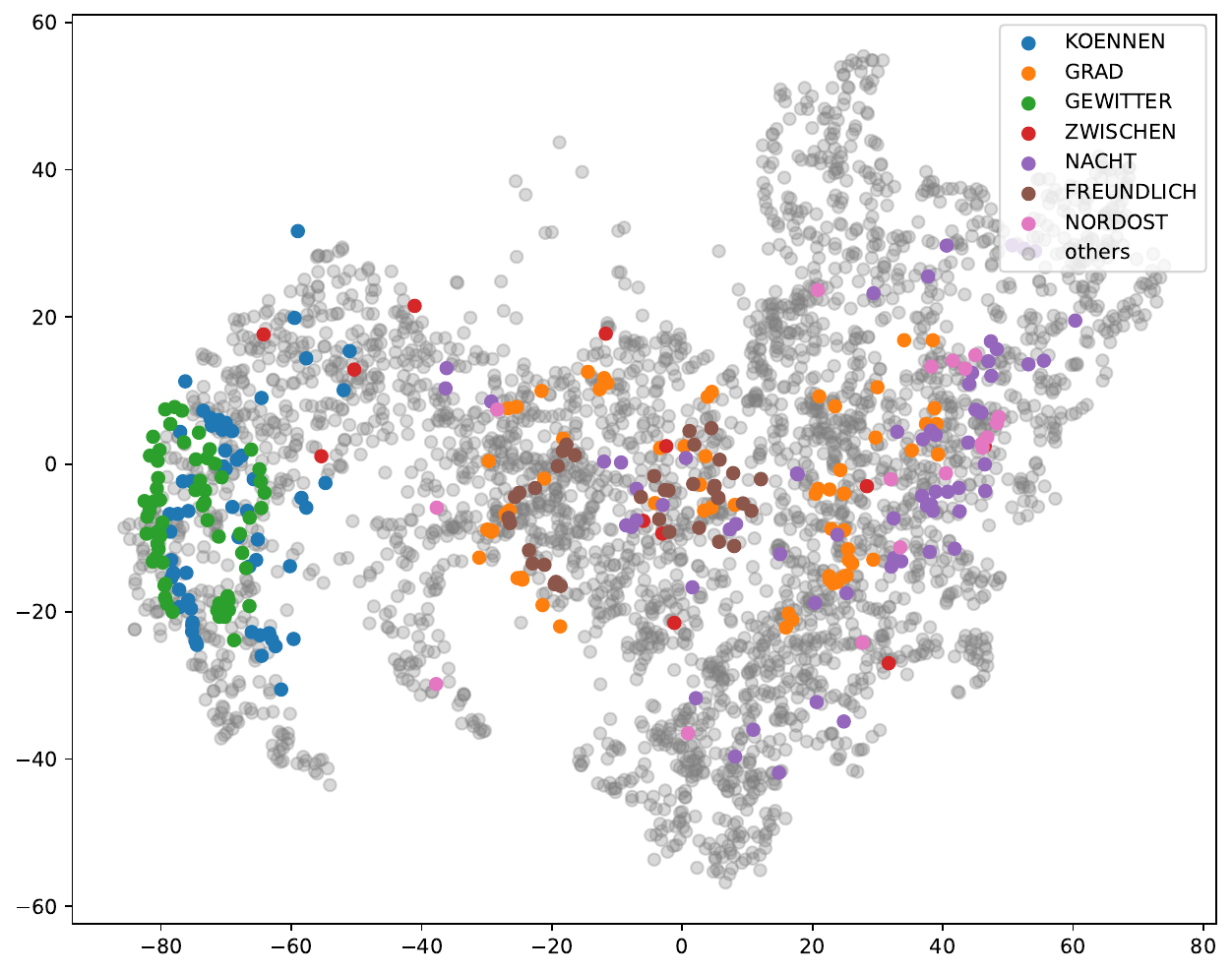}
        \caption{+{\fancyname}(Ours)}
        \label{fig:signcl}
    \end{subfigure}
    \hfill
    \begin{subfigure}{0.19\textwidth}
        \caption*{SDR=74.07\%}  
        \includegraphics[width=\textwidth]{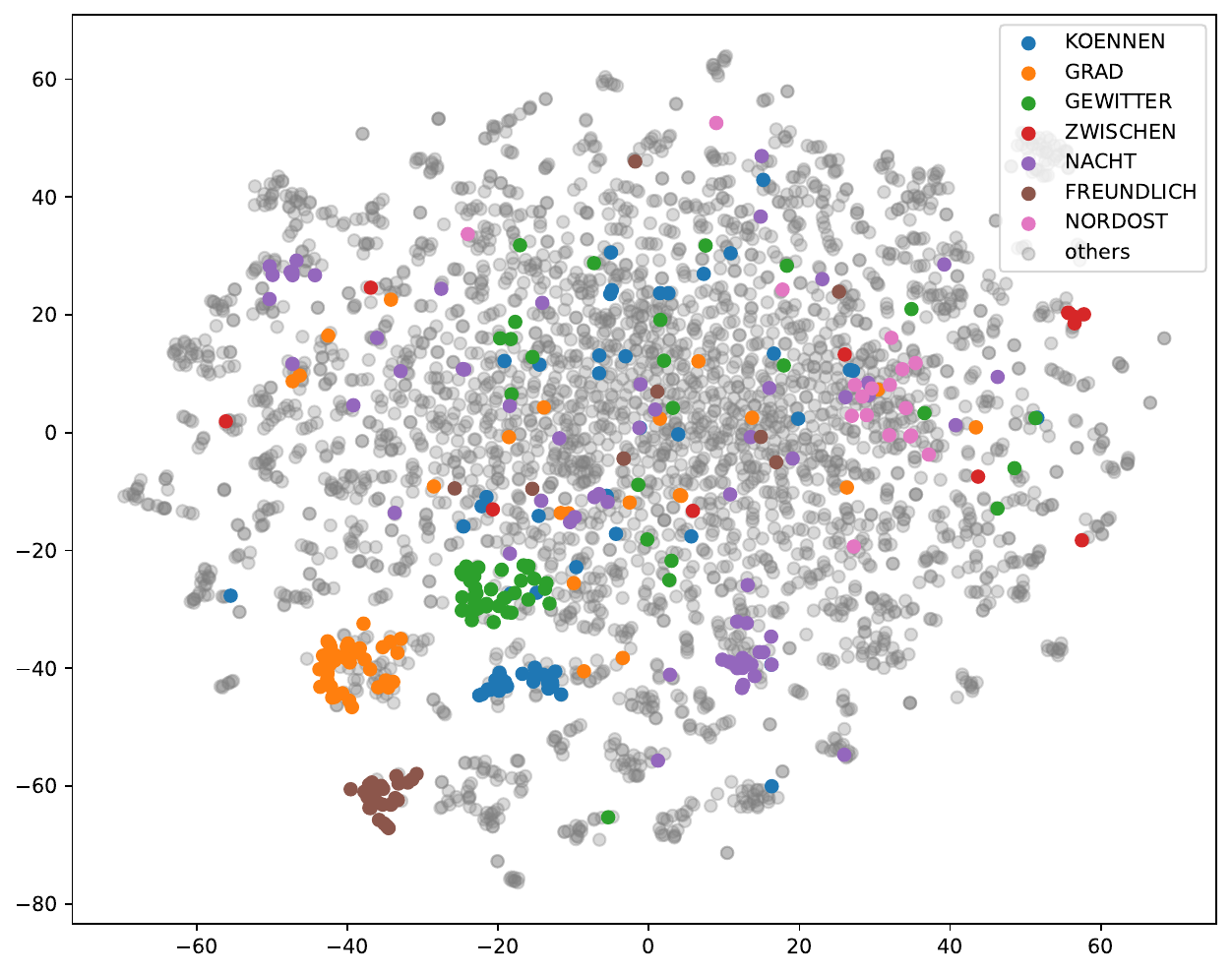}
        \caption{SRP}
        \label{fig:st}
    \end{subfigure}
    \hfill
    \begin{subfigure}{0.19\textwidth}
        \caption*{SDR=66.23\%}
        \includegraphics[width=\textwidth]{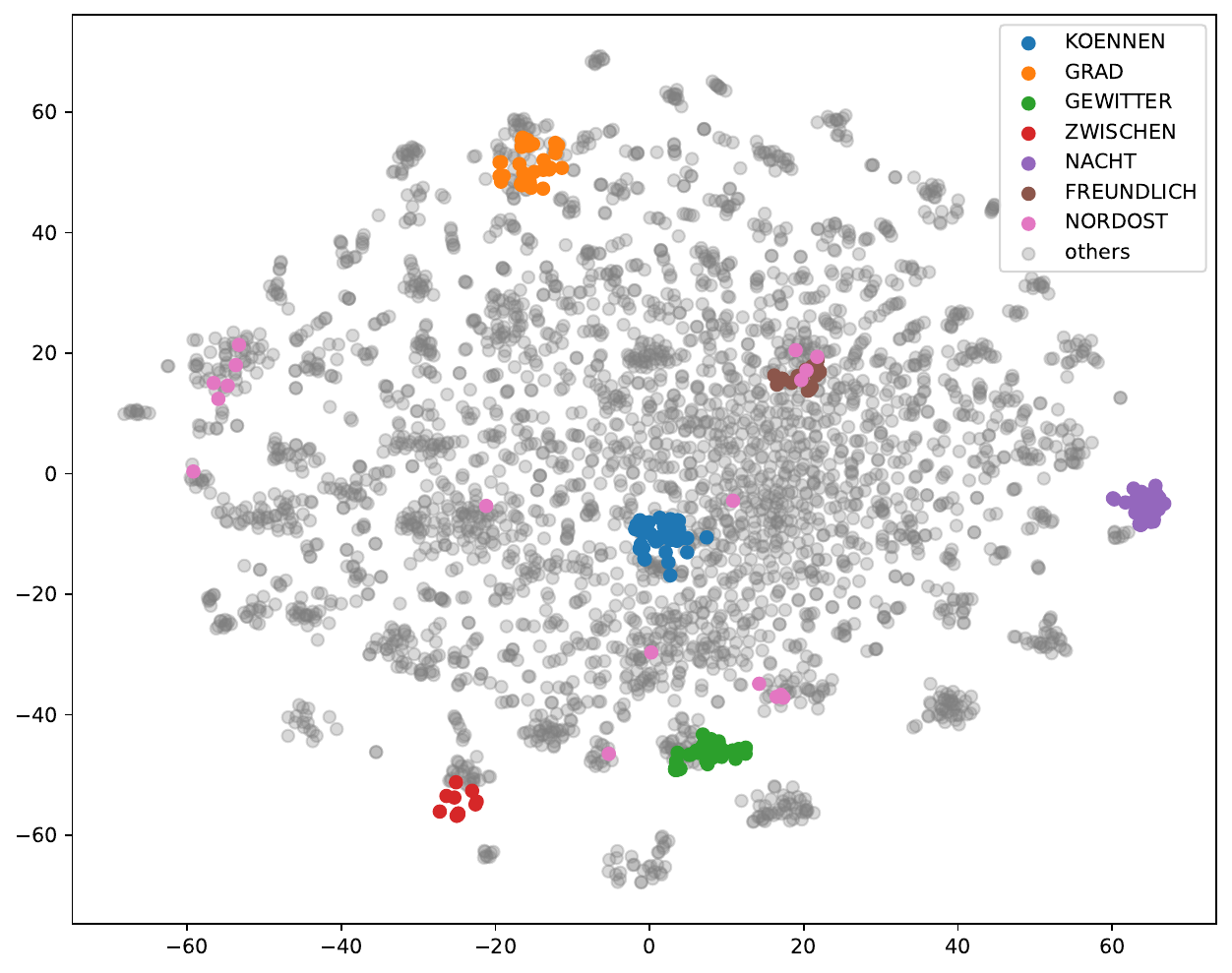}
        \caption{SMKD}
        \label{fig:smkd}
    \end{subfigure}
    \caption{The t-SNE visualization of sign features across existing extraction techniques. SRP, SMKD, and I3D are downloaded from their official websites, while VLP is reproduced with official code. The addition of +{\fancyname} denotes our proposed method that integrates a contrastive learning strategy into the VLP method (see Section~\ref{sec:CL}). Different colors represent sign gestures with distinct semantics. Points in gray represent other sign categories not listed. Better viewed by zooming in.}
    
    \label{fig:sign-embeddings}
\end{figure}

\paragraph{Results and Findings}
\label{sec:cor_results}

Through empirical analysis of various visualized open-source sign features, we have identified a widespread representation density problem across different sign feature extraction methods. As depicted in Figure~\ref{fig:sign-embeddings}, all evaluated methods display a Sign Density Ratio exceeding 50\%, with inevitable overlap of feature representation. Notably, gloss-free methods that do not utilize gloss annotations as additional supervision (e.g., I3D and VLP) exhibit even more severe representation density compared to gloss-based methods. This is evident as sign gestures representing different semantics, indicated by different colors, significantly overlap, resulting in translation ambiguity during inference. Specifically, the Sign Density Ratio (SDR) of VLP is 92.59\%, which is significantly higher than the SDR of SMKD at 66.23\%.

\begin{figure}[htbp]
    \centering
    \begin{subfigure}{0.49\textwidth}
        \caption{Impact on SLR}
        \includegraphics[width=\textwidth]{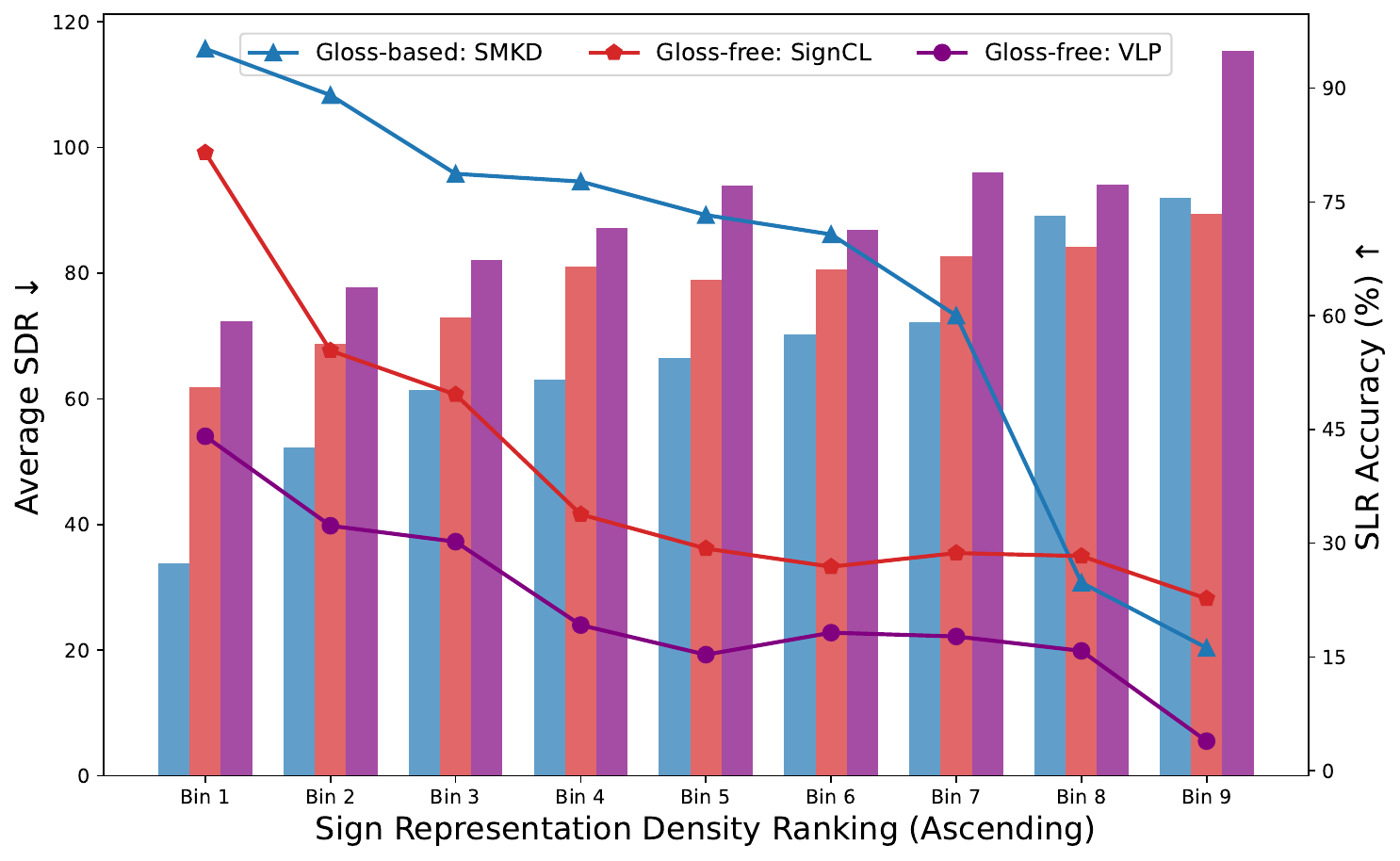}
        \label{fig:SDRvsSLR}
    \end{subfigure}
        \begin{subfigure}{0.48\textwidth}
        \caption{Impact on SLT}
        \includegraphics[width=\textwidth]{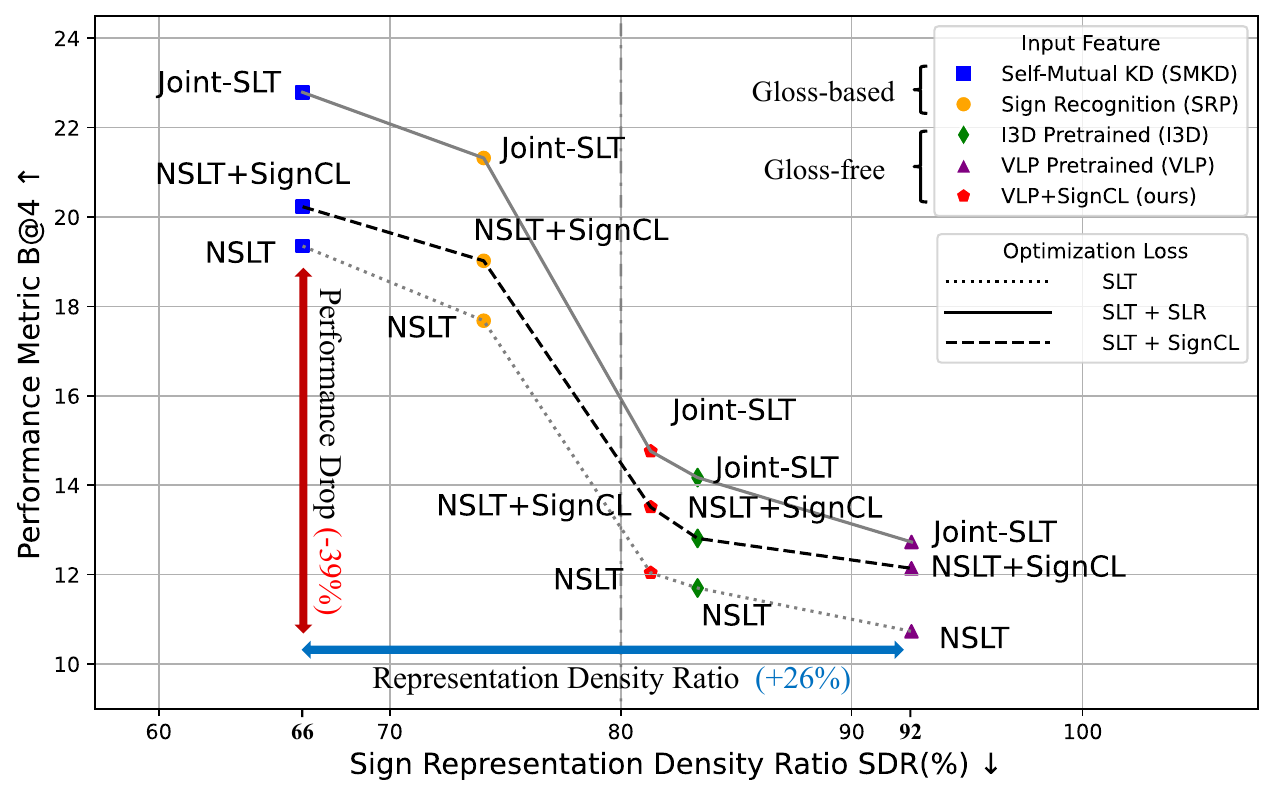}
        \label{fig:SDRvsSLT}
    \end{subfigure}

    \caption{Comparative analysis of representation density and its impact on sign language recognition (SLR) and translation (SLT). The left panel (a) shows the correlation between representation density and SLR accuracy across different sign feature types and sign gesture groups. Binning in this context is based on sorting by gloss density within a group, where higher bins indicate higher density. The right panel (b) illustrates the performance drops in SLT caused by the representation density problem. This figure assesses both the recognition and translation accuracies, reflecting how denser representations impact these metrics.}
    \label{fig:density-acc}
\end{figure}

\subsection{Demonstrating Performance Drop}
\label{sec:ImpactSDR}

This section investigates the impact of representation density on sign language recognition (SLR) and translation (SLT) systems.

\paragraph{General Setups}
This part employs the widely utilized Sign Language Transformer~\cite{camgoz2020sign} (NSLT) as the foundational model for our evaluations.
The choice is because 
its capability to perform both SLT and SLR tasks, as well as take SLR and SLT at the same time (joint-SLT). Additionally, the NSLT framework is well-established within sign language research and benefits from comprehensive documentation and support in open-source sign feature sets and baseline results.
The NSLT relies on sign features derived using a pretrained sign feature extractor. This section studies all types of sign features introduced in Section~\ref{sec:cSDR} to investigate the representation density problem.
We use the Sign Density Ratio (SDR, see Eqn. ~\ref{eq:GDR}) to measure the representation density within each type of input feature. We measure SLR and SLT performance by the recognition accuracy and the BLEU-4~\cite{papineni2002bleu} score (B@4), respectively.

\paragraph{Task Setups}
We set up tasks to examine whether representation density bottlenecks sign language recognition and translation performance.
\begin{itemize}[leftmargin=20pt]
    \item \textbf{Sign Language Recognition}: To evaluate the ability of the extracted sign features to distinguish between different semantic gestures, we use the NSLT~\cite{camgoz2020multi} to perform sign language recognition (SLR) tasks with various types of sign features~\cite{camgoz2018neural} as model input. Due to the limited number of samples for each gesture in the dev set, we rank the sign glosses based on their density using SDR$(G_i)$ under SMKD features (see Eqn. ~\ref{eq:GDR}). These glosses are then divided into nine groups (bins), each containing approximately 60 glosses. The average SDR$(G_i)$ and recognition accuracy for each bin represents the overall density and mean accuracy of the glosses within that bin, respectively.
    
    \item \textbf{Sign Language Translation}: This evaluation aims to demonstrate the impact of representation density on translation tasks. We evaluate various sign features as inputs to the Sign Language Transformer, including SRP, SMKD, I3D, VLP, and VLP+{\fancyname}. These inputs are tested across different translation frameworks, such as NSLT~\cite{camgoz2018neural}, Joint-SLT~\cite{camgoz2020multi}, and NSLT+{\fancyname}. NSLT means use NSLT to perform SLT without CTC loss (gloss-free) and the NSLT+{\fancyname} configuration integrates the proposed contrastive learning strategy into the encoder of NSLT~\cite{camgoz2018neural} models, as detailed in Section~\ref{sec:CL}. 
\end{itemize}

\paragraph{Results and Findings}
As depicted in Figure~\ref{fig:density-acc}, the following observations were made regarding the impact of representation density on both recognition (SLR) and translation (SLT):
\begin{itemize}[leftmargin=20pt]
    \item \textbf{Performance suffers from representation density.} We consistently observed a negative relationship between representation density and performance across all feature types and tasks. Higher representation density leads to worse accuracy in SLR and lower BLEU scores in SLT. Specifically, an increase in the representation density ratio by 26\% can result in a 39\% performance drop in NSLT.
    
    \item \textbf{Gloss-free methods suffer from worse representation density.}
    Gloss-free based feature extractions, which do not use any gloss annotations for assistance (e.g., VLP), typically exhibit higher representation density scores than gloss-based approach (e.g., SDR(VLP)=92.59\% > SDR(SMKD)=66.23\%).
    Using gloss-free features results in worse recognition and translation performance compared to gloss-based feature extractions (e.g., {VLP} vs. {SMKD}).

    \item \textbf{Contrastive learning boosts performance by reducing representation density.} When contrastive learning is applied to augment gloss-free based feature representation learning, i.e., VLP+{\fancyname} for feature extraction or NSLT+{\fancyname} for downstream finetuning, there is a consistent reduction in feature representation density accompanied by a significant improvement in both of the SLR accuracy and the SLT performance (see detail can be found in Section~\ref{sec:CL}). 
\end{itemize}

\section{Contrastive Learning for  Gloss-free Sign Langauge Translation}
\label{sec:CL}
\vspace{-0.4 em}
Contrastive Learning~\cite{jaiswal2020survey}, a popular self-supervised learning algorithm, aims to learn effective representations by pulling positive pairs closer together and pushing negative pairs farther apart.
In this section, we introduce a simple but efficient sign contrastive learning strategy, namely{\fancyname}, which addresses the challenge of the representation density problem in gloss-free sign language translation.

\subsection{Sign Contrastive Learning}
\begin{figure}[htbp]
    \centering
        \begin{subfigure}{0.32\textwidth}
        \includegraphics[width=\textwidth]{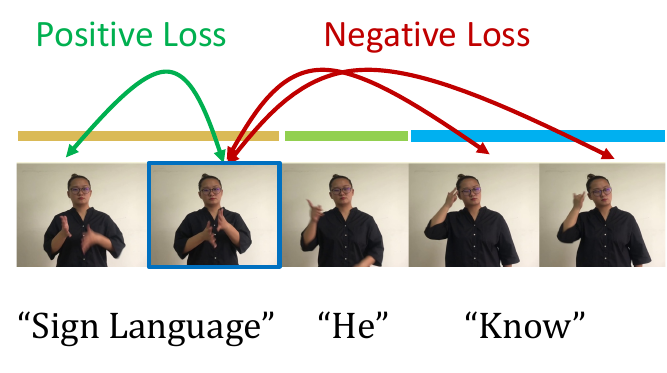}
        \caption{Sign Contrastive Learning}
        \label{fig:scl}
    \end{subfigure}
    \begin{subfigure}{0.32\textwidth}
        \includegraphics[width=\textwidth]{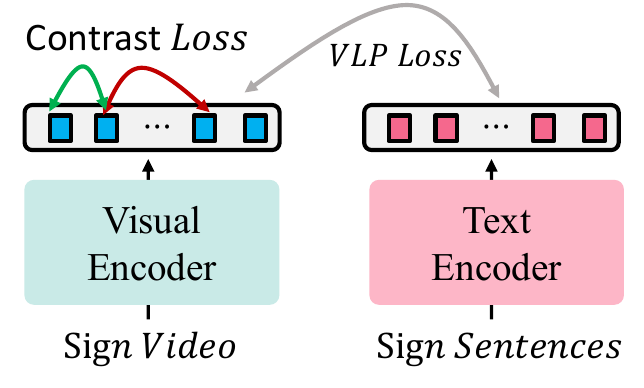}
        \caption{ {\fancyname}  in Pretraining}
        \label{fig:scl_vlp}
    \end{subfigure}
    \begin{subfigure}{0.33\textwidth}
        \includegraphics[width=\textwidth]{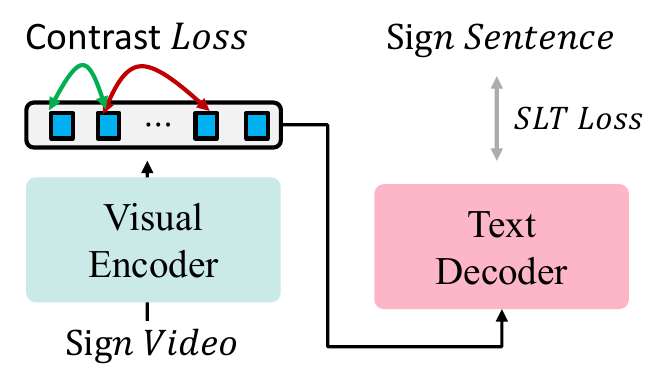}
        \caption{ {\fancyname} in Finetuning}
        \label{fig:scl_slt}
    \end{subfigure}

    \caption{Overview of the {\fancyname}  in gloss-free sign language translation: (a) Sign contrastive learning sampling strategy, (b) Showcases the integration of {\fancyname} in the pretraining stage, and (c) ) Displays the application of {\fancyname} during the finetuning stage.}
    \label{fig:signCL}
\end{figure}

\vspace{-0.4 em}
The key factor in contrastive learning is how to sample positive and negative training pairs. As illustrated in the framework shown in Figure~\ref{fig:scl}, the sampling strategy of {\fancyname} is as follows: if two frames are close enough (e.g., adjacent), they are considered to belong to the same sign gesture and are treated as positive samples. Conversely, if two frames are far apart by double the margin (e.g., $|f_{ed} - f_{st}| > 20$ frames), they are considered to be associated with different semantics and are treated as negative samples. 
Statistically, the average duration of each gesture in sign video is nine frames~\cite{camgoz2018neural,zhou2021improving}, and according to the speech-to-gesture Zipf's Law~\cite{borstell2016distribution}, each gloss represents approximately 2.3 spoken words. Therefore, we set the margin as $ \max(10, \frac{\text{len(frames)}}{\text{len(text)}} \times 2.3)$.

\begin{equation}
\begin{cases}
\text{positive pair } (f_{st}, f_{ed}^+) \text{:} & |f_{ed}^+ - f_{st}| \leq 1 \\
\text{negative pair } (f_{st}, f_{ed}^-) \text{:} & |f_{ed}^- - f_{st}| > 2*margin \\
\end{cases}; 
\end{equation}

\vspace{-0.4 em}
\begin{equation}
\mathcal{L}_{SignCL}= \frac{1}{N} \sum_{{st}=1}^{N} \left[  \text{d}(f_{st}, f_{ed}^+) + \max(0, m - \text{d}(f_{st}, f_{ed}^-)) \right];
\end{equation}

Where \( d \) is the distance function, i.e., Euclidean distance for frame features \( (f_{st}, f_{ed}) \), and $N$ is the total number of frames in one sign video, \( N = \text{len(frames)} \). The margin parameter 
$m$ is used to prevent the features of the negative pair from being too far away. We empirically set \( m = 64 \) based on the average Inter-Gloss Distance (see Eqn. \ref{eq:Inter}) of gloss-based sign features (e.g., SMKD\cite{hao2021self}).

\vspace{-1 em}
\subsection{Integrating Contrastive Learning into Sign Language Translation Tranining}

As illustrated in Figures~\ref{fig:scl_vlp} and~\ref{fig:scl_slt}, \textit{\fancyname} can be integrated into both the sign feature extraction pretraining stage (e.g., Visual-Language Pretraining~\cite{zhou2023gloss}) and the downstream task finetuning stage (e.g., GFSLT-VLP~\cite{zhou2023gloss}). The optimization objective for these approaches is the weighted sum of $\mathcal{L}_{\fancyname}$ and the original objective loss (e.g., VLP Loss for pretraining and SLT loss for finetuning~\cite{camgoz2018neural, zhou2023gloss}), defined as:
\begin{align}
\mathcal{L}= \lambda*{\mathcal{L}}_{\fancyname} + {\mathcal{L}}_{MLE};
\label{eqn:loss_total}
\end{align}

Where $L_{MLE}$ is the original objective loss in the pertaining or finetuning.

\vspace{-0.3 em}
\section{Experiments}
\vspace{-0.6 em}
\label{sec:main_exps}
In this Section, we conduct experiments to demonstrate the efficiency of proposed {\fancyname} in reducing representation density and boosting gloss-free sign language translation performance. 
Specifically, we apply {\fancyname} to the Sign Language Transformer~\cite{camgoz2020multi} to facilitate a direct comparison with prior empirical analyses of the representation density problem in Section~\ref{sec:ImpactSDR}. 
Additionally, we integrate {\fancyname} into the GFSLT-VLP~\cite{zhou2023gloss} framework, a robust new gloss-free baseline that improves SLT through pretraining and finetuning.

\vspace{-0.6 em}
\subsection{Experiments on Sign Language Transformer}
\label{sec:pilot_ex_SLT}
In Section~\ref{sec:signDensity}, we investigate the representation density problem using the Sign Language Transformer (SLT)~\cite{camgoz2020multi} and the PHOENIX-2014T~\cite{camgoz2018neural} and CSL-Daliy~\cite{zhou2021improving} Dataset. These benchmarks are chosen for their established relevance in sign language translation research, including gloss-based and gloss-free based.
Here, we first conduct experiments on the same framework and dataset to facilitate direct comparison with the prior empirical analyses.

\textbf{Experiment Settings:} 
In this experiment, we introduce \fancyname\ as additional supervision information in the encoder of SLT under gloss-free settings. This enhanced model is referred to as {+\fancyname}.

\textbf{Results and Findings:}
The integration of {\fancyname} into the SLT has significantly improved translation performance across all test conditions by reducing the representation density, as shown in Table~\ref{tab:compare_slt}. Notably, {\fancyname} encourages SLT to learn a more distinct feature distribution, reducing the Sign Density Ratio (SDR) significantly, e.g., 66.23 to 62.18 and 92.59 to 81.30.

Figures~\ref{fig:SDRvsSLR} and~\ref{fig:SDRvsSLT} show experiments on SLR and SLT tasks using features with varying SDRs as inputs to SLT. The representation density reduction leads to observable improvements in both recognition accuracy (red line vs. purple line in Figure~\ref{fig:SDRvsSLR}) and translation BLEU score (purple point vs. red point in Figure~\ref{fig:SDRvsSLT}). Further details and additional experiment results on the CSL-Daily dataset are provided in Appendix~\ref{apd:reslut_cls}.

Table~\ref{tab:compare_slt} presents a comparative analysis of representation density and performance on the PHOENIX-2014T dataset. The inclusion of {\fancyname} during VLP feature extraction or SLT training processes significantly enhances performance metrics. WERs (Word Error Rates) in the gloss-free set, derived from an independent SLR task, are specifically used to probe the quality of sign features and do not participate in the SLT training process. This analysis underscores the significant enhancements brought by {\fancyname} in terms of both efficiency and effectiveness in SLT frameworks.

\begin{table*}[ht!]
    \centering
    \begin{tabular}{l  | ccc |ccc }
        \toprule
       \multirow{2}{*}{\bf Feature Type} &  \multicolumn{3}{c|}{\bf PHOENIX-2014T }  &  \multicolumn{3}{c}{\bf  CSL-Daily } \\
        & \textbf{${SDR}$ $\downarrow$}  & WER$\downarrow$ & B@4$\uparrow$ & \textbf{${SDR}$ $\downarrow$} & WER$\downarrow$ & B@4$\uparrow$\\
        \midrule
        \multicolumn{7}{c}{\cellcolor{gray!30}\centering\emph{\phantom{\hspace{16.2em} }\em Gloss-based}} \\
        Joint-SLT~\cite{camgoz2020sign}  / Self-Mutual KD~\cite{min2021visual} &  {66.23} & {25.38} &  {22.79}  &{48.34}&{29.52} &{11.61} \\
        ~~+  {\fancyname} into Feature Extraction & 62.18  &24.76 &23.23&- & - & - \\
        ~~+  {\fancyname} into Finetuning & 66.23  &25.12 &22.92&- &-  &- \\

        ~~+  {\fancyname} into both & \textbf{62.18}  & \textbf{24.58} & \textbf{23.46} &- & -  &- \\
        \multicolumn{7}{c}{\cellcolor{gray!30}\centering\emph{\phantom{\hspace{16.2em} }\em Gloss-free}} \\

        SLT~\cite{camgoz2018neural}  / VLP Pretrained~\cite{zhou2023gloss} &  {92.59} & \textit{69.72}  &  {10.73}  &{76.55}& \textit{85.78} &{1.82} \\
        ~~+  {\fancyname} into Feature Extraction &81.30  & \textit{63.33} & 12.04  & 68.39  & \textit{80.71} &2.15 \\
        ~~+  {\fancyname} into Finetuning & 92.59  & -  & 12.14& 76.55 &-  &2.19 \\

        ~~+  {\fancyname} into both & \textbf{81.30}  & - & \textbf{13.51} & \textbf{68.39}  & -  & \textbf{2.53} \\

        \bottomrule
    \end{tabular}
    \caption{Comparative analysis of representation density and performance on the PHOENIX-2014T dataset. "+{\fancyname}" indicates the inclusion of the proposed contrastive learning strategy during VLP (Video Language Processing) feature extraction or SLT (Sign Language Translation) training processes. WERs (Word Error Rates) in the gloss-free set are derived from an independent SLR (Sign Language Recognition) task, used specifically for probing the quality of sign features. These WERs do not participate in the SLT training process.
}
    \label{tab:compare_slt}
\end{table*}

\subsection{Experiments on Gloss-free Sign Language Translation}

Gloss-free sign language translation, which does not rely on gloss annotations, has become a trend as it makes the approach more generalizable.
In the realm of gloss-free sign language translation, GFSLT-VLP~\cite{zhou2023gloss} stands out as a strong new baseline. It incorporates CLIP~\cite{radford2021learning} and MBART~\cite{chipman2022mbart} for model pretraining and finetuning. In this set of experiments, we use GFSLT-VLP as the baseline model and integrate the proposed {\fancyname} into the framework to demonstrate the effectiveness of our method in both pretraining and finetuning settings.

\begin{table*}[th!]
    \centering
    \setlength{\tabcolsep}{3pt}
    \begin{tabular}{l  | c  | ccccc}
        \toprule
       \multirow{2}{*}{\bf Model} &  
      \multirow{1}{*}{\bf Density }  &  \multicolumn{5}{c}{\bf Performance} \\  
          & $SDR\downarrow$&  R@L$\uparrow$ & B@1 $\uparrow$ & B@2 $\uparrow$& B@3 $\uparrow$& B@4 $\uparrow$\\
        \midrule
        NSLT~\cite{camgoz2018neural,camgoz2020multi}  & - & 30.07& 29.86 & 17.52 &11.96 &  9.00 \\
        GASLT~\cite{yin2023gloss}  &- &39.86 &39.07 & 26.74 &21.86& 15.74 \\
        GFSLT~\cite{zhou2023gloss} &- & 40.93&  41.39 & 31.00 & 24.20 &  19.66\\
        GFSLT-VLP~\cite{zhou2023gloss} & - &42.49 & 43.71& 33.18 & 26.11  & 
 21.44 \\
        Sign2GPT(w/PGP)~\cite{wong2024signgpt} & -& 48.90 & 49.54& 35.96 & 28.83& 22.52 \\
        \midrule
        GFSLT-VLP~\cite{zhou2023gloss} & 68.53 &42.97  & 42.13 & 32.04 &25.62   & 
 21.25 \\
        ~~~~+  {\fancyname} into Pretraining  & 62.68 &\textbf{49.25} &\textbf{49.99} &36.73&29.76& 22.69 \\
        ~~~~+  {\fancyname} into Finetuning & 62.73 & 48.17 & 48.56 &35.04 & 27.73& 22.16 \\
       ~~~~+  {\fancyname} into Two State & \textbf{62.32} & 49.04 & 49.76 & \textbf{36.85}  & \textbf{29.97}  & \textbf{22.74}  \\
        \hline
        Improvement & \textcolor{red}{-6.21} & \textcolor{blue}{+6.07} & \textcolor{blue}{+7.63} & \textcolor{blue}{+4.81} & \textcolor{blue}{+4.35} & \textcolor{blue}{+1.49 }\\
        \bottomrule
    \end{tabular}
    \caption{Improvement in the GFSLT-VLP framework by reducing representation density on PHOENIX-2014T test set. 
    "+{\fancyname} into Pretraining" indicates applying the proposed contrastive learning strategy during the pretraining stage, while  "+{\fancyname} into Finetuning" indicates the inclusion of the {\fancyname} during the finetuning stage. "+{\fancyname} into Two State" means plus {\fancyname} both in pertaining and finetuning states.}
    \label{tab:main_dsl}
\end{table*}

\begin{table*}[th!]
    \centering
    \begin{tabular}{l  | c  | ccccc}
        \toprule
       \multirow{2}{*}{\bf Model} &  
      \multicolumn{1}{c|}{\bf Density}  &  \multicolumn{5}{c}{\bf Performance} \\  
           & $SDR\downarrow$& R@L$\uparrow$ & B@1 $\uparrow$ & B@2 $\uparrow$& B@3 $\uparrow$& B@4 $\uparrow$\\
    \midrule
        GASLT~\cite{yin2023gloss}  &   -   & 20.35  &19.90&9.94&5.98& 4.07 \\
        NSLT~\cite{camgoz2018neural,camgoz2020multi}  &   -   & 34.54  &34.16&19.57&7.56&  7.56 \\
        GFSLT~\cite{zhou2023gloss} &  -  &35.16   &37.69 &23.28& 14.93&  9.88 \\
        GFSLT-VLP~\cite{zhou2023gloss}   &-& 36.44 &39.37 & 24.93& 16.26  & 
 11.00 \\
        Sign2GPT(w/PGP)~\cite{wong2024signgpt} & -& 42.36& 41.75&28.73 &20.60 & 15.40 \\
        \midrule
        GFSLT-VLP~\cite{zhou2023gloss} &  58.20  &39.08 &36.37 &23.32&15.45& 11.10 \\
        ~~~~+  {\fancyname} into Pretraining & 55.24 &47.38 &46.20 &32.33&22.35& 15.85  \\
        ~~~~+  {\fancyname} into Finetuning & 55.03  &48.26 &46.53&32.41&22.42& 15.98 \\
        ~~~~+  {\fancyname} into Two States & \textbf{54.61}  &\textbf{48.92} &\textbf{47.47}&\textbf{32.53}&\textbf{22.62}& \textbf{16.16}\\
        \hline
        Improvement & \textcolor{red}{-3.59} & \textcolor{blue}{+9.84} & \textcolor{blue}{+11.10} & \textcolor{blue}{+9.21} & \textcolor{blue}{+7.17} & \textcolor{blue}{+5.06} \\
        \bottomrule
    \end{tabular}
     \caption{ Enhancing GFSLT-VLP by reducing representation density on CSL-Daily test set.}
    \label{tab:main_cls}
    \vspace{-0.6 em}
\end{table*}

\textbf{Experiment Settings: }
\label{sec:main_ex_setup}
This set of experiments is conducted using the PHOENIX-2014T~\cite{camgoz2018neural} and CSL-Daily~\cite{zhou2021improving} datasets. We reproduce GFSLT-VLP using the official code and integrate {\fancyname} into both the pretraining and finetuning stages. All models and training details are consistent with those used in GFSLT-VLP~\cite{zhou2023gloss}, with the sole exception being the incorporation of {\fancyname}, weighted by $\lambda=0.01$, as illustrated in Figure~\ref{fig:sign-embeddings} and Equation~\ref{eqn:loss_total}. Further details are provided in Appendix~\ref{apd:hype_para}.

\textbf{Results and Findings: }
\label{sec:main_ex_results}
Tables~\ref{tab:main_dsl} and~\ref{tab:main_cls} compare our proposed methods with existing gloss-free sign language translation approaches. The results demonstrate that integrating the proposed {\fancyname} strategy into the GFSLT-VLP framework consistently reduces representation density and significantly boosts translation performance, whether {\fancyname} is applied during pretraining, finetuning, or both stages. Specifically, compared to the baseline model GFSLT-VLP ~\cite{zhou2023gloss}, our approach achieves a substantial improvement of 45.58\% (+5.06) in the BLEU-4 score on the CSL-Daily dataset, without any increase in the number of parameters. Additionally, despite having significantly fewer parameters ($\sim$600M vs. $\sim$1.7B), our approach achieves better performance than Sign2GPT~\cite{wong2024signgpt}, which leverages large-scale pretrained vision and language models for sign language translation.

\subsection{Qualitative Analysis}

\begin{figure}[ht!]
\centering
\begin{minipage}{\textwidth}
    \centering
    \includegraphics[width=0.85\textwidth]{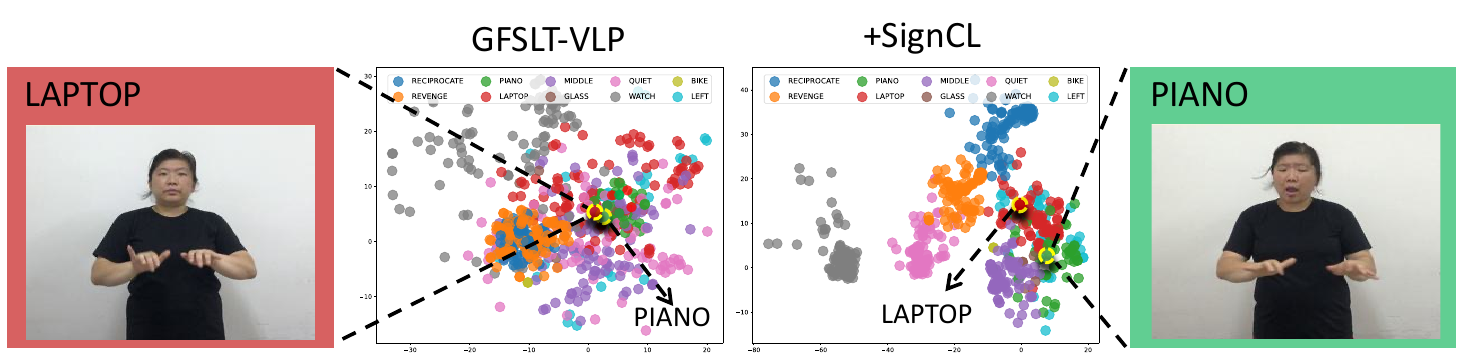}
\end{minipage}

\begin{minipage}{\textwidth}
    \centering
    \begin{tabular}{l|l}
    \midrule
    Reference & \begin{CJK}{UTF8}{gkai}妈妈给我买\colorbox{green!30}{钢琴}。 \end{CJK} (Mom bought me a \colorbox{green!30}{piano}.) \\
    GFSLT-VLP & \begin{CJK}{UTF8}{gkai}妈妈给我买\colorbox{red!30}{笔记本电脑}。(Mom bought me a \colorbox{red!30}{laptop}.) \end{CJK}\\
    GFSLT-VLP + {\fancyname} & \begin{CJK}{UTF8}{gkai}妈妈给我买\colorbox{green!30}{钢琴}。 \end{CJK} (Mom bought me a \colorbox{green!30}{piano}.) \\
    \midrule
    \end{tabular}
\end{minipage}

\caption{Qualitative comparison of translation results on CSL-Daily test set. The {red background} denotes model misinterpretations about the sign gestures, while {green} one means accurate recognition. Content in ( ... ) is English translation for non-Chinese readers.}
\label{fig:case_motivation}
\end{figure}

\begin{CJK}{UTF8}{gkai}
To understand our \fancyname\ approach in scenarios of addressing representation density, we present a case from the CSL-Daily dataset in Figure~\ref{fig:case_motivation}. As shown, the way to display sign gestures for “电脑” (laptop) and “钢琴” (piano) differ subtly. 
As indicated by the t-SNE results, the representations of these two semantically different gestures are closely packed together in the feature space, causing the baseline GFSLT-VLP model to incorrectly translate “钢琴” (piano) as “电脑” (laptop). In contrast, our proposed {\fancyname} effectively separates the representations of “电脑” (laptop) and “钢琴” (piano) in the feature space, enabling the accurate translation of “钢琴” (piano).
\end{CJK}

\section{Conclusion}

In this work, we identify a crucial representation density problem in gloss-free sign language translation. Our systematic investigation reveals that this problem persists across various existing sign feature extraction methods and causes sharp performance drops in both sign language recognition and translation, particularly in gloss-free methods. To address this problem, we propose a simple but effective contrastive learning strategy, termed {\fancyname}. Our experiments demonstrate that {\fancyname} encourages gloss-free models to learn more discriminative features and significantly reduces representation density. Furthermore, our experiments show that {\fancyname} improves translation performance across various frameworks and datasets by a significant margin, achieving a new state-of-the-art in gloss-free sign language translation. 
We illustrate the effectiveness of {\fancyname} through detailed examples in our qualitative analysis. Finally, we provide several ablation studies for a better understanding of {\fancyname} and discuss the limitations and potential societal impacts of this work in the Appendix~\ref{sec:APD}.

\section{Limitations}
\label{apd:limitaiton}

Our work, while promising, has several limitations that should be considered:

\textbf{Boundary Cases:} We assume that adjacent frames output the same sign gestures, while distant frames belong to different sign gestures. This assumption might not hold in special sign language videos with extensive repetitive gestures. In extreme cases, {\fancyname} might affect feature convergence.

\textbf{Semantic Similarity}: {\fancyname} does not account for the semantic similarity between sign gestures, which can result in increased feature distances between semantically similar gestures. This could potentially affect the learning of linguistic features.

Despite acknowledging these limitations, our experiments demonstrate that our approach works effectively in most cases. We will address these issues in future work to further enhance the robustness and applicability of our method.

\section{Acknowledgement}
This work was supported in part by the National Natural Science Foundation of China (Grant No. 92370204), in part by the National Key R\&D Program of China (Grant No. 2023YFF0725001), in part by the Guangzhou-HKUST (GZ) Joint Funding Program (Grant No. 2023A03J0008), and in part by the Education Bureau of Guangzhou Municipality.
This work was also supported by the Guangzhou Municipal Science and Technology Project (No. 2024A04J4390).

\clearpage
{\small

\bibliographystyle{apalike}  
\bibliographystyle{plain}
\bibliography{egbib}
}


\appendix
\newpage
\clearpage

\section{Appendix}
\label{sec:APD}
\subsection{Hyper-parameters of Baselines}
\label{apd:hype_para}

\textbf{Sign Language Transformers Baseline: }
Table~\ref{tab:hype_para} presents the hyper-parameters of Sign Language Transformers used in this work.

\begin{table}[h!] 
    \centering
     \setlength{\tabcolsep}{1pt} 
    \begin{tabular}{l c c c}
        \toprule
        \multirow{2}{*}{\bf Parameter} & \multirow{2}{*}{\bf PHOENIX-2014T} &  \multirow{2}{*}{\bf CSL-Daily} \\
        \\
        \midrule
         encoder-layers  & 3  & 1 \\
         decoder-layers  & 3  & 1 \\
         attention heads & 8 & 8 \\
         ctc-layers & 1 & 1 \\
         hidden size & 512 & 512 \\
         activation function & gelu & gelu \\
         learning rate   & $1 \cdot 10^{-3}$ & $1 \cdot 10^{-3}$ \\
         Adam $\beta$    &  (0.9, 0.98) &  (0.9, 0.98)\\
         label-smoothing & 0.1  & 0.1 \\
         max output length & 30 & 50 \\
         dropout         & 0.3  & 0.3 \\
         batch-size      & 128 & 128 \\
        \bottomrule
    \end{tabular}
    \caption{Hyperparameters of Sign Language Transformer models.}
    \label{tab:hype_para}
\end{table}

\textbf{Gloss-Free Sign Language Translation  Baseline: }
\label{sec:GFSLT}
The Gloss-Free Sign Language Translation (GFSLT) model incorporates various modules designed for processing sign language input without the use of glosses. Below is the detailed architecture used in this work:

\begin{table}[h!]
    \centering
    \begin{tabular}{lccc}
        \toprule
        \textbf{Module} & \textbf{Stride} & \textbf{Kernel} & \textbf{Output Size} \\
        \midrule
        Sign Input & - & - & $B \times T \times 224 \times 224 \times 3$ \\
        Resnet w/o fc & - & - & $B \times T \times 512$ \\
        Conv1D-BN1D-RELU & 1 & 5 & $B \times T \times 1024$ \\
        MaxPooling1D & 2 & 2 & $B \times \frac{T}{2} \times 1024$ \\
        Conv1D-BN1D-RELU & 1 & 5 & $B \times \frac{T}{2} \times 1024$ \\
        MaxPooling1D & 2 & 2 & $B \times \frac{T}{4} \times 1024$ \\
        Linear-BN1D-RELU & - & - & $B \times \frac{T}{4} \times 1024$ \\
        Transformer Encoder & - & - & $B \times \frac{T}{4} \times U$ \\
        \midrule
        Text Input & - & - & $B \times U$ \\
        Word Embedding & - & - & $B \times U \times 1024$ \\
        Transformer Decoder & - & - & $B \times U \times 1024$ \\
        FC & - & - & $B \times U \times C$ \\
        \bottomrule
    \end{tabular}
    \caption{Detailed Gloss-Free SLT (GFSLT) Framework. $B$ represents batch size, $T$ denotes the length of the longest input sign video in the batch, and $U$ is the length of the longest input text in the batch. It is copied from GFSLT-VLP~\cite{zhou2023gloss}.}
    \label{tab:GFSLT-framework}
\end{table}

\noindent
\subsection{Parameter Sensitivity Analysis of the {\fancyname}}

\subsubsection{Sensitivity Analysis on Dynamically Estimated Margin}
The margin for negative sampling dynamically depends on the estimated average margin of each gloss, calculated as \( \text{len(frames)} / \text{len(text)} \times \text{speech-to-gesture Zipf’s factor} \), with a minimum threshold set at 10. The Zipf’s factor, set at 2.3, refers to the speech-to-gesture application of Zipf’s Law.

\begin{figure}[t!]
    \centering
    \includegraphics[width=0.65\textwidth]{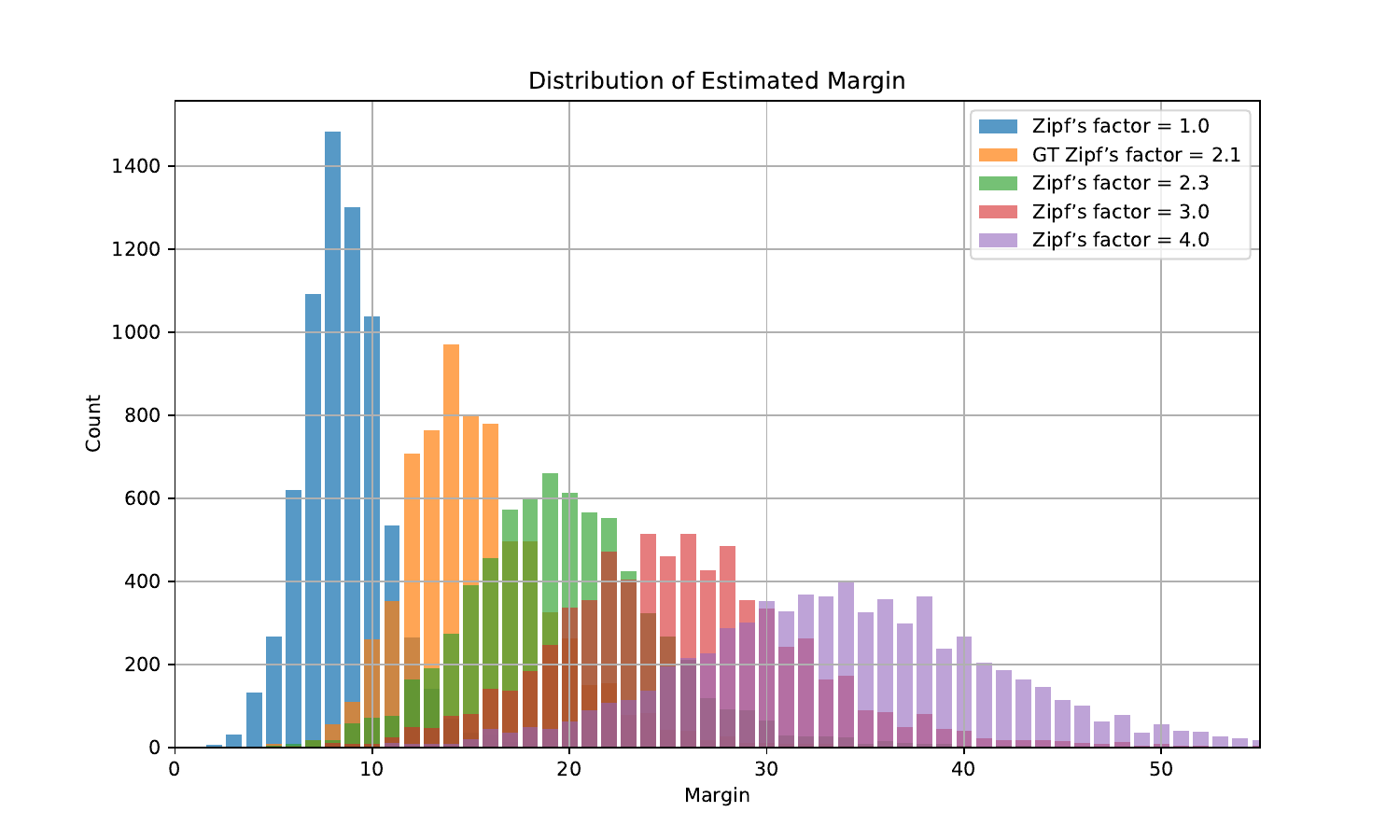}
    \caption{ The distribution of the estimated margin during training on the PHOENIX-2014T dataset.
    The green distribution represents our current paper’s method (factor = 2.3), while the orange distribution shows the ground truth calculated based on gloss annotations.}
    \label{fig:margin}
\end{figure}

We calculated the distribution of the dynamically estimated margin, with the results displayed in the table below. A more detailed distribution can be seen in Figure~\ref{fig:margin}.

\noindent
\textbf{Experiment Setup:}
To conduct a principled analysis, we evaluated the threshold values at \([0, 10, 20, 30, 40, 50]\). Here, a threshold of 0 indicates that the margin is dominated by the dynamically estimated margin, while a threshold of 50 suggests dominance by the fixed threshold.

\noindent
\textbf{Experiment Results:}
We uniformly trained for 80 epochs on the PHOENIX-2014T dataset due to resource limitations. The results, as shown in the table below, indicate that SignCL is not sensitive to the threshold parameter, with a variance of 0.062.

\begin{table}[ht!]
\centering
\begin{tabular}{l | c c c c c c}
\hline
\textbf{Threshold} & 0 & 10 & 20 & 30 & 40 & 50 \\
\hline
B@4 & 17.24 & 17.63 & 17.55 & 17.63 & 17.13 & 17.11 \\
\hline
\end{tabular}
\caption{Threshold sensitivity analysis results.}
\label{tab:threshold_sensitivity}
\end{table}

\begin{table}[ht!]
\centering
\begin{tabular}{l | c c c c c}
\hline
\textbf{Zipf’s factor} & 1 & GT & 2.3 & 3 & 4 \\
\hline
B@4 & 17.45 & 17.89 & 17.63 & 17.29 & 16.26 \\
\hline
\end{tabular}
\caption{Sensitivity to Zipf's factor.}
\label{tab:zipfs_factor_sensitivity}
\end{table}

\subsubsection{Sensitivity Analysis on integrating {\fancyname} into the SLT framework. }
As shown in Eqn. ~\ref{eqn:loss_total}, we vary the hyperparameter \(\lambda\) over the range \([10^{-3}, 10^{-2}, 10^{-1}, 10^{0}, 10^{1}]\) and conduct repeated experiments on the PHOENIX-2014T dataset with GFSLT-VLP.

\begin{figure}[t!]
\centering
\includegraphics[width=0.65\textwidth]{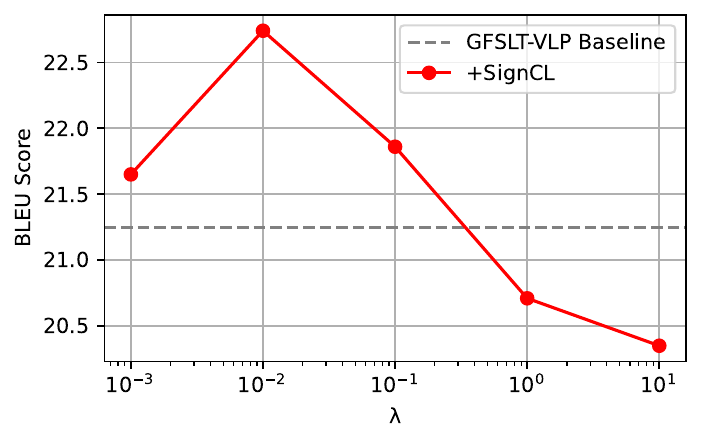}
\caption{The effect of the hyperparameter 
\(\lambda\) on BLEU scores. 
the grey dashed line indicates the baseline performance of GFSLT-VLP, while the red solid line represents the performance with {\fancyname} integrated.}
\label{fig:lambda}
\end{figure}

As shown in Figure~\ref{fig:lambda}, excessively incorporating {\fancyname} into the model can negatively impact the SLT task. Empirically,   we find that  \(\lambda = 10^{-2}\) achieves a balance between reducing representation density and improving translation performance.

\subsection{Ablation Studies}
\label{apd:ablation}
We conduct ablation studies to investigate the impact of different loss components in the +{\fancyname} approach during both the pretraining and fine-tuning stages. 
It is copied from Tabel~\ref{tab:main_dsl}
, but with an ablation perspective.

\begin{table*}[th!]
    \centering
    \begin{tabular}{cc | cc  | c | cc}
        \toprule
        \multicolumn{2}{c|}{\bf Pretraining Stage} & \multicolumn{2}{c|}{\bf Finetuning Stage} & 
        \multicolumn{1}{c|}{\bf Density} & \multicolumn{2}{c}{\bf Performance} \\
         \textbf{VLP Loss} &
         \textbf{{\fancyname} Loss} &  
         \textbf{SLT Loss} & 
         \textbf{{\fancyname} Loss} & 
        $SDR\downarrow$ & R@L$\uparrow$ & B@4$\uparrow$
        \\
        \midrule
        \midrule
        
        \xmark & \xmark &  \cmark & \xmark & 72.83 & 38.67 & 18.53 \\
        \xmark & \cmark &   \cmark & \xmark & 63.23 & 39.12 & {18.71}\\
        \cmark & \xmark &   \cmark & \xmark & 68.53 & 42.97 & 21.25\\
        \cmark & \cmark &   \cmark & \xmark & 62.68 & 49.25 & 22.69 \\
        \midrule
        \xmark & \xmark &  \cmark & \cmark & 69.54 & 41.78 & 19.81 \\
        \xmark & \cmark &   \cmark & \cmark & 63.67& 44.52& {20.03}\\
        \cmark & \xmark &   \cmark & \cmark & 62.73 & 48.17 & 22.16\\
        \cmark & \cmark &   \cmark & \cmark & \textbf{62.32} & \textbf{49.23} & \textbf{22.74}\\
        \bottomrule
    \end{tabular}
     \caption{Ablation study on the impact of different loss components in the +{\fancyname} approach.}
    \label{tab:abl_loss}
\end{table*}

\subsection{{Correlation between Representation Density and Recognition Performance}}

\subsubsection{The efficiency of sign-gloss mapping building up }
To calculate SDR, we need to establish the mapping relationship between input frames and gloss categories. This section presents the performance of our trained gloss-sign aligner. The experimental methodology follows the approach outlined in XmDA~\cite{ye-etal-2023-cross}.

\begin{table}[htb]
    \centering
    \setlength{\tabcolsep}{1.5pt} 
    \begin{tabular}{l ccc}
        \toprule
       \multirow{2}{*}{\bf Dataset} &  \multicolumn{3}{c}{\bf \textbf{WER}$\downarrow$ }  \\
        \cmidrule{2-4}
        & Train & {Test} & Dev \\
        \midrule
        PHOENIX-2014T & 8.68 & 8.03 & 25.28  \\
        CSL-Daily & 9.23 & 8.39 & 29.32 \\
        \bottomrule
    \end{tabular}
     \caption{Evaluation of the gloss-sign aligner effectiveness and generalizability with WER (\%) (the lower the better).
}
    \label{tab:s2gAligner}
\end{table}

\subsubsection{Correlation Coefficient}

We analyze the relationship between the representation density of individual glosses and their recognition accuracy using the Sign Language Transformer on the PHOENIX-2014T dataset, leveraging the Self-Mutual Knowledge Distillation (SMKD) method for feature extraction.
We compute the following correlation coefficients:

\textbf{Pearson Correlation Coefficient~\cite{cohen2009pearson}:} This measures the linear relationship between recognition accuracy \(Acc(G_i)\) for gloss \(G_i\) and the density metric $SDR(G_i)$, calculated as:

\begin{equation}
\label{eq:r}
r = \frac{\sum (Acc(G_i) - \bar{Acc})(SDR(G_i) - \bar{SDR})}{\sqrt{\sum (Acc(G_i) - \bar{Acc})^2 \sum (SDR(G_i) - \bar{SDR})^2}}
\end{equation}

\textbf{Spearman's Rank Correlation Coefficient~\cite{sedgwick2014spearman}:} This assesses the monotonic relationship between two datasets by considering the rank order of values.

As shown in Table~\ref{tab:g2gAligner}, both correlation coefficients indicate a medium negative correlation between the Sign Density Ratio (\(SDR\)) and sign recognition performance (Acc). 
This suggests that higher representation density correlates with poorer recognition performance.
Specifically, Inter-Gloss Distance (\(D^{inter}_{G_i}\))  shows a strong positive correlation, meaning that greater distances between different glosses correlate with better recognition performance. 
All Spearman P-values are lower than 0.01, confirming the high confidence in the non-randomness of these correlations.

\begin{table*}[!ht]
    \centering
    \begin{tabular}{l | ccc | ccc}
        \toprule
        \multirow{2}{*}{\bf Correlation / Acc} & \multicolumn{3}{c|}{\bf PHOENIX-2014T} & \multicolumn{3}{c}{\bf CSL-Daily} \\
        & { \(D^{inter}_{G_i}\uparrow\)} & {\(D^{intra}_{G_i}\downarrow\)} & {SDR $\downarrow$} & { \(D^{inter}_{G_i}\uparrow\)} & {\(D^{intra}_{G_i}\downarrow\)} & {SDR $\downarrow$} \\
        \midrule
        \textbf{Pearson \(r\)} & 0.36 & -0.22 & -0.35 & 0.30 & -0.14 & -0.20 \\
        \textbf{Spearman \(\rho\)} & 0.43 & -0.24 & -0.34 & 0.45 & -0.16 & -0.22 \\
        \textbf{P-value} & 2.6E-17 & 5.5E-6 & 4.7E-11 & 6.6E-19 & 3.0E-3 & -2.7E-5\\
        \bottomrule
    \end{tabular}
    \caption{Correlation analysis between sign recognition performance and density metrics.}
    \label{tab:g2gAligner}
\end{table*}

\subsubsection{More Experiment Results on Sign Language Transformer}
\label{apd:reslut_cls}
In this section, we present additional experimental results using the Sign Language Transformer (NSLT) on the CSL-Daily dataset to further validate the effectiveness of the proposed {\fancyname} strategy. We compare various feature extraction methods to assess their representation density and translation performance.

\begin{table*}[ht!]
    \centering
    \begin{tabular}{l  | c  | cc cc }
        \toprule
       \multirow{2}{*}{\bf Feature Type} &  \multicolumn{1}{c}{\bf Density } &  \multicolumn{4}{|c}{\bf Performance} \\
        & \textbf{${SDR}$ $\downarrow$}  & SLR(WER $\downarrow$) & Joint-SLT & NSLT & +{\fancyname}(ours)\\
        \midrule
        \multicolumn{6}{c}{\cellcolor{gray!30}\centering\emph{\phantom{\hspace{8.2em} }\em Gloss-based Feature Extraction}} \\
        Sign Recognition~\cite{camgoz2020sign} & 74.07  & 29.59 & 21.32 &17.68&19.02\\
        Self-Mutual KD~\cite{min2021visual} &  \textbf{66.23} & \textbf{25.38} &  \textbf{22.79}  &\textbf{19.35}&\textbf{20.23}\\
        \multicolumn{6}{c}{\cellcolor{gray!30}\centering\emph{\phantom{\hspace{8.2em} }\em Gloss-free Feature Extraction}} \\
        I3D Pretrained~\cite{yin2023gloss} & 83.33 & 61.74 & 14.17 & 11.70 &12.81\\
        VLP Pretrained~\cite{zhou2023gloss} & 92.59 & 69.72  & 12.73 & 10.73 &12.14\\
        + {\fancyname} (ours) & \textbf{81.30} & \textbf{63.33} & \textbf{14.76}  &  \textbf{12.04} &\textbf{13.51}\\
        \bottomrule
    \end{tabular}
    \caption{Comparative analysis of representation density and performance on the PHOENIX-2014T dataset. "+{\fancyname} (ours)" indicates the inclusion of the proposed contrastive learning strategy during VLP feature extraction or NSLT training processing.}
    \label{tab:compare}
\end{table*}

\begin{table*}[ht!]
    \centering
    \begin{tabular}{l  | c  | cc cc }
        \toprule
       \multirow{2}{*}{\bf Feature Type} &  \multicolumn{1}{c}{\bf Density } &  \multicolumn{4}{|c}{\bf Performance} \\
        & \textbf{${SDR}$ $\downarrow$}  & SLR(WER $\downarrow$) & Joint-SLT & NSLT & +{\fancyname}(ours)\\
        \midrule
        \multicolumn{6}{c}{\cellcolor{gray!30}\centering\emph{\phantom{\hspace{8.2em} }\em Gloss-based Feature Extraction}} \\
        Self-Mutual KD~\cite{min2021visual} &  \textbf{48.34} & \textbf{29.52} &  \textbf{11.61}  &\textbf{8.97}&\textbf{10.35}\\
        \multicolumn{6}{c}{\cellcolor{gray!30}\centering\emph{\phantom{\hspace{8.2em} }\em Gloss-free Feature Extraction}} \\
        VLP Pretrained~\cite{zhou2023gloss} & 76.55 & 85.78  & 2.93 & 1.82 &2.19\\
        + {\fancyname} (ours) & \textbf{68.39} & \textbf{80.71} & \textbf{3.18}  &  \textbf{2.29} &\textbf{2.53}\\
        \bottomrule
    \end{tabular}
    \caption{Comparative analysis of representation density and performance on the CSL-Daily dataset. The Self-Mutual KD features are provided by XmDA~\cite{ye-etal-2023-cross} and the VLP feature is reproduced with official code. Due to the incomplete open source of the CSL-Daily dataset, we were unable to obtain features for Sign Recognition and I3D Pretraining.}

    \label{tab:compare_csl}
\end{table*}






\subsection{Broader Impacts}
\label{sec:Broader}

This paper focuses on research in sign language translation, which has the potential to significantly benefit individuals who are deaf or hard of hearing. By improving the accuracy and efficiency of sign language translation, our work can facilitate better communication between individuals with hearing impairments and the broader community. This can help break down communication barriers, promoting inclusivity and equal opportunities in various social, educational, and professional settings.

\label{sec:appendix}



\newpage
\section*{NeurIPS Paper Checklist}

\begin{enumerate}

\item {\bf Claims}
    \item[] Question: Do the main claims made in the abstract and introduction accurately reflect the paper's contributions and scope?
    \item[] Answer: \answerYes{} 
    \item[] Justification: The abstract and introduction accurately reflect the main contributions and scope of the paper. 

\item {\bf Limitations}
    \item[] Question: Does the paper discuss the limitations of the work performed by the authors?
    \item[] Answer: \answerYes{} 
    \item[] Justification:  In the Appendix, we discuss the limitations of this work.

\item {\bf Theory Assumptions and Proofs}
    \item[] Question: For each theoretical result, does the paper provide the full set of assumptions and a complete (and correct) proof?
    \item[] Answer: \answerNA{} 
    \item[] Justification: The paper does not include theoretical results. 

    \item {\bf Experimental Result Reproducibility}
    \item[] Question: Does the paper fully disclose all the information needed to reproduce the main experimental results of the paper to the extent that it affects the main claims and/or conclusions of the paper (regardless of whether the code and data are provided or not)?
    \item[] Answer: \answerYes{} 
    \item[] Justification: We disclose all the information needed to reproduce the main experimental results in the Methodology, Experiment, and Appendix sections. We will release the code upon acceptance of the paper. 

\item {\bf Open access to data and code}
    \item[] Question: Does the paper provide open access to the data and code, with sufficient instructions to faithfully reproduce the main experimental results, as described in supplemental material?
    \item[] Answer: \answerYes{} 
    \item[] Justification: We commit to releasing the code and models upon acceptance of the paper. All the data and baselines are based on open-source benchmarks. 

\item {\bf Experimental Setting/Details}
    \item[] Question: Does the paper specify all the training and test details (e.g., data splits, hyperparameters, how they were chosen, type of optimizer, etc.) necessary to understand the results?
    \item[] Answer: \answerYes{} 
    \item[] Justification: We provide all the training and test details necessary to understand the results in the Methodology, Experiment, and Appendix sections. 

\item {\bf Experiment Statistical Significance}
    \item[] Question: Does the paper report error bars suitably and correctly defined or other appropriate information about the statistical significance of the experiments?
    \item[] Answer: \answerNo{} 
    \item[]  Justification: Error bars are not reported because performing multiple runs for each experiment would be too computationally expensive. 

\item {\bf Experiments Compute Resources}
    \item[] Question: For each experiment, does the paper provide sufficient information on the computer resources (type of compute workers, memory, time of execution) needed to reproduce the experiments?
    \item[] Answer: \answerYes{}  
    \item[] Justification: All experiments are conducted using PyTorch on 8*NVIDIA A800 GPUs for about 12 hours. 
    
    
\item {\bf Code Of Ethics}
    \item[] Question: Does the research conducted in the paper conform, in every respect, with the NeurIPS Code of Ethics \url{https://neurips.cc/public/EthicsGuidelines}?
    \item[] Answer:\answerYes{} 
    \item[] Justification: We adhere to the NeurIPS Code of Ethics, since the paper does not include any content or practices that violate ethical guidelines.
    

\item {\bf Broader Impacts}
    \item[] Question: Does the paper discuss both potential positive societal impacts and negative societal impacts of the work performed?
    \item[] Answer: \answerYes{} 
    \item[] Justification: We discuss the potential societal impacts of the paper in the Appendix. 
    
\item {\bf Safeguards}
    \item[] Question: Does the paper describe safeguards that have been put in place for responsible release of data or models that have a high risk for misuse (e.g., pretrained language models, image generators, or scraped datasets)?
    \item[] Answer: \answerNA{} 
    \item[] Justification: The paper does not involve data or models that have a high risk for misuse. 

\item {\bf Licenses for existing assets}
    \item[] Question: Are the creators or original owners of assets (e.g., code, data, models), used in the paper, properly credited and are the license and terms of use explicitly mentioned and properly respected?
    \item[] Answer: \answerYes{} 
    \item[] Justification: The paper properly credits the creators or original owners of all used assets, and properly respects the license and terms of use. 

\item {\bf New Assets}
    \item[] Question: Are new assets introduced in the paper well documented and is the documentation provided alongside the assets?
    \item[] Answer: \answerNA{} 
    \item[] Justification: No new assets are introduced in the paper.

\item {\bf Crowdsourcing and Research with Human Subjects}
    \item[] Question: For crowdsourcing experiments and research with human subjects, does the paper include the full text of instructions given to participants and screenshots, if applicable, as well as details about compensation (if any)? 
    \item[] Answer: \answerNA{} 
    \item[] Justification: The paper does not involve crowdsourcing nor research with human subjects. 

\item {\bf Institutional Review Board (IRB) Approvals or Equivalent for Research with Human Subjects}
    \item[] Question: Does the paper describe potential risks incurred by study participants, whether such risks were disclosed to the subjects, and whether Institutional Review Board (IRB) approvals (or an equivalent approval/review based on the requirements of your country or institution) were obtained?
    \item[] Answer: \answerNA{} 
    \item[] Justification: The paper does not involve crowdsourcing nor research with human subjects, thus there are no study participants, no risks to disclose to subjects, and no need for Institutional Review Board (IRB) approvals.

\end{enumerate}

\end{document}